\documentclass[letterpaper]{article} 
\usepackage{aaai25}  
\usepackage{times}  
\usepackage{helvet}  
\usepackage{courier}  
\usepackage[hyphens]{url}  
\usepackage{graphicx} 
\usepackage{natbib}  
\usepackage{caption} 
\usepackage{algorithm}
\usepackage{algorithmic}

\usepackage{newfloat}
\usepackage{listings}
\DeclareCaptionStyle{ruled}{labelfont=normalfont,labelsep=colon,strut=off} 
\floatstyle{ruled}
\newfloat{listing}{tb}{lst}{}
\floatname{listing}{Listing}
\title{HurriCast: Synthetic Tropical Cyclone Track Generation for Hurricane Forecasting}
\author{
    Shouwei Gao\textsuperscript{1},
    Meiyan Gao\textsuperscript{2},
    Yuepeng Li\textsuperscript{2},
    Wenqian Dong\textsuperscript{1}
}

\affiliations{
    \textsuperscript{1}Oregon State University, Corvallis, OR, USA \\
    \textsuperscript{2}Florida International University, Miami, FL, USA \\
    gaosho@oregonstate.edu, megao@fiu.edu, yuepli@fiu.edu, wenqian.dong@oregonstate.edu
}

\usepackage{bibentry}

\begin{document}

\maketitle

\begin{abstract}
The generation of synthetic tropical cyclone(TC) tracks for risk assessment is a critical application of preparedness for the impacts of climate change and disaster relief, particularly in North America. Insurance companies use these synthetic tracks to estimate the potential risks and financial impacts of future TCs. For governments and policymakers, understanding the potential impacts of TCs helps in developing effective emergency response strategies, updating building codes, and prioritizing investments in resilience and mitigation projects. In this study, many hypothetical but plausible TC scenarios are created based on historical TC data HURDAT2 (HURricane DATA 2nd generation). A hybrid methodology, combining the ARIMA and K-MEANS  methods with Autoencoder, is employed to capture better historical TC behaviors and project future trajectories and intensities. It demonstrates an efficient and reliable in the field of climate modeling and risk assessment. By effectively capturing past hurricane patterns and providing detailed future projections, this approach not only validates the reliability of this method but also offers crucial insights for a range of applications, from disaster preparedness and emergency management to insurance risk analysis and policy formulation.
\end{abstract}

%

\section{Introduction}

Many coastal U.S. communities are highly susceptible to significant hurricane and flood damage from sea level rise, high tides, storm surges, and extreme rainfall due to dense populations, high property values, and disproportionately vulnerable populations in low-lying areas. 
The frequency of billion-dollar weather-related disasters is rising steadily~\cite{noaa2018}. Severe hazards, including tornadoes, hail, and wind, caused over $32B$ of property damage in 2017~\cite{noaa2018,insurancejournal2016}, and wind and tornadoes have killed an average of 160 people annually in the last 10 years~\cite{nws2018fatalities}. 
In urban centers such as South Florida, this is not a future phenomenon that will be triggered by, for example, future sea level rise, but rather already a common occurrence that is generally referred to as “nuisance or recurrent flooding”~\cite{valle2017spatial,wdowinski2016increasing}. For instance, recently, one neighborhood in the Florida Keys was flooded for almost three months just by tides— an unprecedented event in that locality that was not predicted. Much of this increased flooding of heavily urbanized areas is attributable to recent sea level rise and changing ocean dynamics~\cite{ezer2013sea,ezer2014accelerated}.

Artificial intelligence (AI) and machine learning (ML) have been utilized by diverse environmental science user groups to revolutionize the understanding and prediction of high-impact atmospheric and ocean science phenomena and to create new educational pathways to develop a larger and more diverse AI/ML and environmental science workforce.
For example, prior work~\cite{gagne2019interpretable, gagne2017storm, lagerquist2019deep, mcgovern2017using} demonstrated that AI techniques can skillfully predict convective hazards~\cite{board2016next}. 
Testing in the National Oceanic and Atmospheric Administration (NOAA)’s Hazardous Weather Testbed ~\cite{clark2012overview} demonstrated the need for trustworthy AI.
AI/ML techniques are useful and promising as they can effectively identify, through data mining, complex relationships between external drivers and urban flooding using observational datasets. 
By employing reverse engineering and automatic learning methodologies it is often possible that AI/ML is able to solve complex, unstructured problems with a fraction of the computing power and execution time required by traditional direct and first-principle methods. 
A board agent demonstrates the importance of AI/ML techniques in environmental science. NOAA and DOE have identified AI as a high priority in their new strategic AI plan~\cite{noaa2019strategies}, within Objective 4.3: "Support the National Artificial Intelligence Research Institutes Program with NSF by collaborating with appropriate institutes on AI."



\begin{figure*}[!ht]
  \centering
  \includegraphics[width=0.8\textwidth]{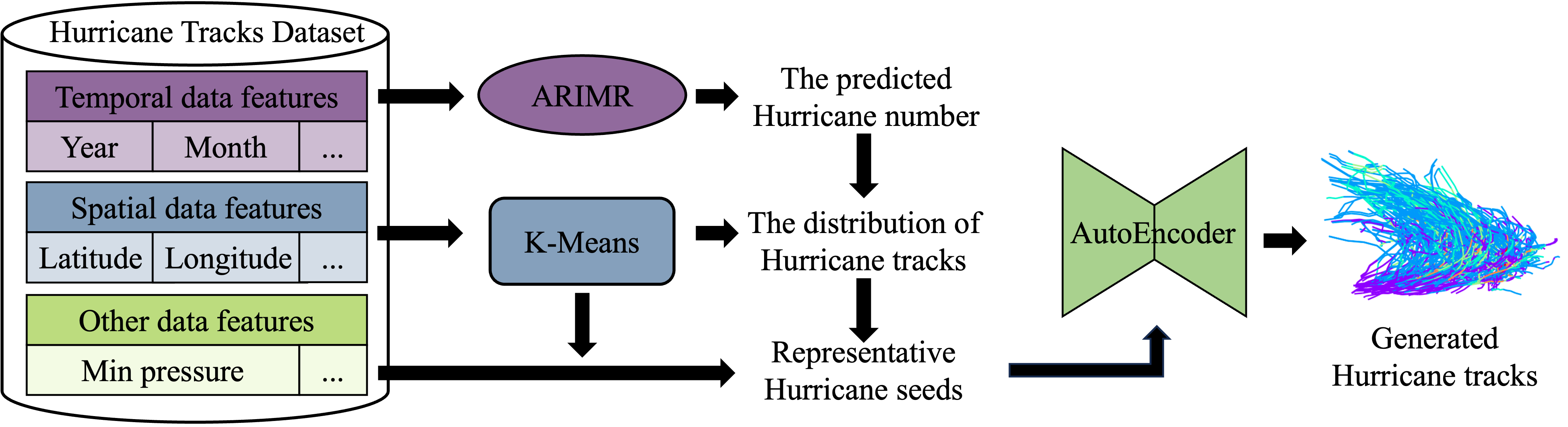}
  \caption{Flowchart of the hurricane track simulator} 
  \label{fig:framework} 
\end{figure*}

Despite this critical importance, the ability to simulate the tropical cyclone’s state at the scales needed to understand and manage these uses and their impacts is not readily accessible to key stakeholder communities.
Several barriers inhibit acquiring real-world or appropriate synthetic datasets, processing these data, and efficiently incorporating data from different phenomena and scales for tropical cyclone system research. 
%
%
Firstly, there is a clear shortage in the hurricane data we have.
When modeling hurricanes, we look at three main areas: hazard, exposure, and vulnerability. Each area needs its own set of data. But often, this data is incomplete, inconsistent, or even missing~\cite{SurfaceP}. Furthermore, the current best datasets~\cite{fang2011smart} of hurricanes cover around 100 years, and even within this, only some of the data is considered trustworthy and complete, as mentioned in \cite{powell2005state}.

Secondly, existing approaches~\cite{AMultivariateS, powell2005state} often overlook comprehensive temporal and spatial analyses before generating synthetic data for tropical cyclone trajectories. This oversight means that they may not fully account for changes in hurricane behavior over time or across different locations. Understanding the patterns of hurricanes across various time frames (e.g., seasons) and geographical regions is crucial for accurate predictions.
For example, the trajectory of a hurricane in a coastal region may differ significantly from one that forms further out at sea (detailed in section 3.6). Similarly, hurricanes in consecutive years may share more characteristics than those from decades apart. However, SOTA methods ~\cite{WEI2023105398, Moradi2016, Mudigonda2017SegmentingAT, bose2022simulation} often overlook temporal and spatial variations, risking the loss of critical information and leading to inaccurate or unreliable predictions.


To address these challenges, we begin by analyzing both the temporal and spatial tendencies of tropical cyclone tracks. We use the ARIMA model to estimate future tropical cyclone track counts based on temporal trends and apply K-MEANS clustering to group and sample tropical cyclone tracks with similar spatial characteristics. After that, we pick the most representative tropical cyclone tracks as seeds and use an Autoencoder to simulate hurricane patterns derived from these seeds. 
We make the following contributions in this paper:
\begin{itemize}
\item We explore the feasibility of combining statistical modeling and data-driven learning to simulate tropical cyclone tracks for risk prediction. Specifically, we adopt statistical models (i.e., the ARIMA model and K-MEANS) as the pre-processing step to extract input data for an Autoencoder.

\item We introduce a framework -- HurriCast, an automated yet lightweight software tool for generating synthetic tropical cyclone tracks for specific regions. These synthetic tracks are particularly valuable for stakeholders who have sparse datasets, enabling more comprehensive analysis and improved decision-making.

\item We demonstrate the effectiveness of HurriCast by validating it against real-world hurricane data, showing that it can accurately reproduce key features of tropical cyclone behavior. This validation underscores the utility of our approach in augmenting existing datasets and enhancing the reliability of hurricane risk assessments.

\end{itemize}

\begin{figure*}[htbp]
  \centering
  \includegraphics[width=0.95\textwidth]{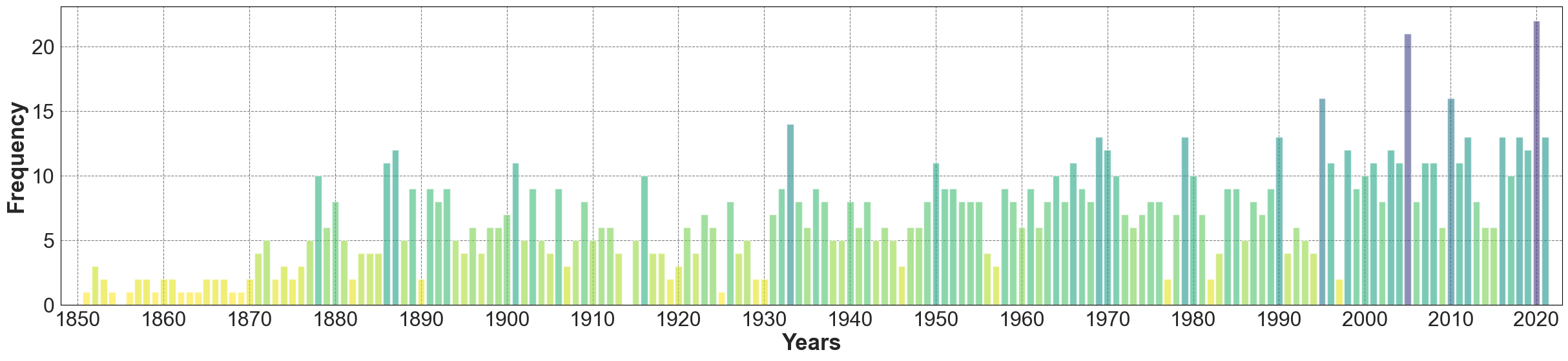}
  \caption{Hurricane frequency across different years} 
  \label{fig:yearananlysis} 
\end{figure*}

\section{Background and Related Work}


Hurricanes, powerful tropical cyclones characterized by intense winds and heavy rainfall, are phenomena that can inflict immense damage on communities, ecosystems, and economies. Predicting the track (i.e., path and intensity) of these storms has long been a crucial aspect of meteorological research and public safety efforts.
Most hurricanes originate from specific regions, like the Western North Atlantic or the Eastern North Pacific. Once formed, hurricanes tend to follow general tracks/patterns based on prevailing winds, oceanic conditions, and other atmospheric factors. Hurricanes can make landfall, causing massive destruction, or dissipate over water. The location and timing of landfall can vary widely, making predictions challenging.

The most accurate way to forecast future hurricane events is using numerical methods, which consists of the following steps:
(1) create an atmospheric model using a grid system for the designated region;
(2) set initial conditions based on data from satellites, weather balloons, and ocean buoys;
(3) incorporate important weather variables to make short-term predictions about atmospheric changes;
(4) examine the interactions between the atmosphere and ocean to estimate the hurricane's path and strength;
(5) use ensemble forecasting, which involves running several simulations with varied starting conditions, to refine and broaden the forecast's accuracy.
This approach, however, is impractical as the computation of each hurricane track can be extremely time-consuming. 

Instead, the synthesis of hurricane tracks is a lightweight process of generating potential future hurricane tracks based on historical and current data. The objective is to prepare for scenarios that, while not having occurred historically, are plausible given known patterns and conditions. These synthetic tracks provide an enriched dataset that aids in comprehensive risk assessment, especially in regions with limited historical hurricane data. There are several ways to generate the hurricane tracks.

\noindent\textbf{Statistical Methods.} Traditional methods mainly involve statistical models that rely on historical hurricane track data. These models take into account the initial conditions and current atmospheric state and then extrapolate potential future pathways based on historical patterns. Many studies have utilized statistical and dynamical climate models (\cite{Goerss2000}, \cite{Knutson2013} ) to model North Atlantic tropical cyclone activity and how they are affected by different baseline climate conditions. 
Study~\cite{nakamura2021early} proposed a climate-conditioned simulation of North Atlantic tropical storm tracks to assess early-season hurricane risk and considered both the influence of the large-scale climate conditions and the historical tropical storm data. Although directly using the past hurricane datasets is appealing, statistical forecast methods still have a poor performance with respect to dynamic models\cite{GiffardRoisin2018DeepLF}.

\noindent\textbf{Stochastic Methods. }Some synthesis methods use stochastic or probabilistic models. These models introduce an element of randomness, ensuring a wide variety of potential tracks, thus broadening the scope of risk assessments\cite{Weinstein2021, Pinelli2020}. Goerss et al.\cite{Goerss2000} used a dynamical climate model to model North Atlantic TC activity(including occurrence, landfall rates and etc.) and changes according to the baseline conditions. However, these methods are in low resolution and expensive to run repeatedly\cite{SGCMIntegrations, tellusa14705}.

\noindent\textbf{Machine Learning Methods.} Recent advancements in technology have introduced machine learning and deep learning methods for hurricane track prediction. These methods use complex algorithms trained on vast amounts of data to predict hurricane tracks, potentially uncovering novel patterns and relationships not apparent in traditional models. 
A model designed by Moradi et al.\cite{Moradi2016} uses a sparse recurrent neural network from only track data for trajectory prediction of Atlantic hurricanes. Mudigonda et al. \cite{Mudigonda2017SegmentingAT} designed a hybrid ConvNet-LSTM network to learn the $(x, y)$ trajectory coordinates and showed their results on the two-dimensional fixed scale map.  In Bose et al. study\cite{bose2022simulation}, a deep learning approach was used to simulate Atlantic hurricane tracks and features, achieving better accuracy compared to traditional statistical methods. However, these studies can not generate hurricane tracks based on the temporal data to give several years of hurricane risk assessment which is interested by insurance industry.

In this research, we champion a hybrid strategy: commencing with the pre-processing of historical data via statistical methods to derive representative tracks; Subsequently, stochastic noise is incorporated into a machine learning framework, enabling perturbations to authentically emulate hurricane track patterns.

\begin{figure}[h]
  \centering
  \includegraphics[width=.43\textwidth]{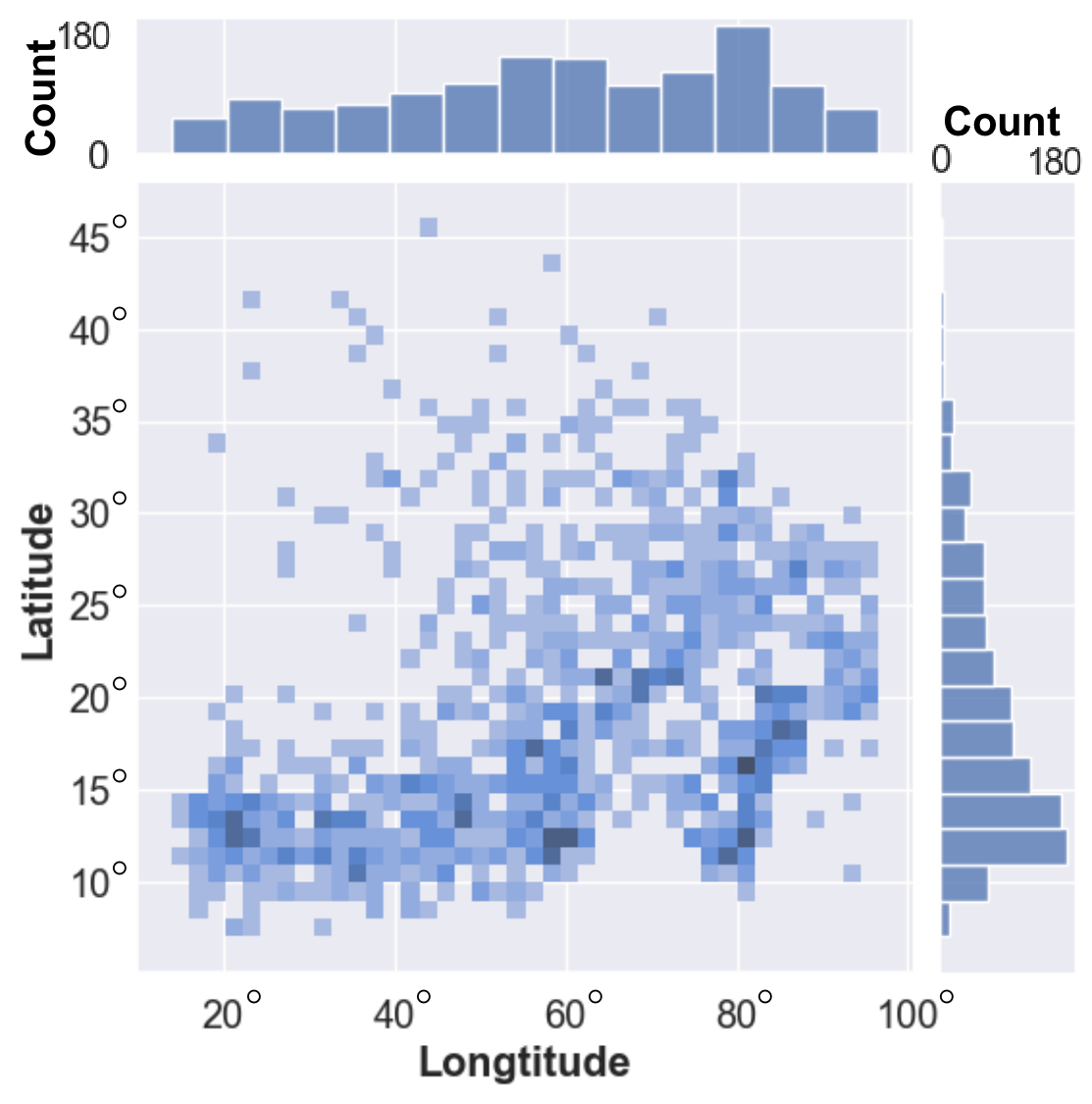}
  \caption{Joint probability of hurricane frequency by longitude and latitude (grid size: 5$^\circ$)} 
  \label{fig:jointprobability} 
\end{figure}

\section{Methods}

\subsection{Design Overview}

The workflow of HurriCast, illustrated in Figure~\ref{fig:framework}, is designed to be user-friendly and fully automated. Users only need to provide the historical hurricane track data for the target region. HurriCast then analyzes this data to forecast future hurricane coverage and intensity. 

HurriCast employs a hybrid method that integrates statistical models with data-driving methods to conduct both temporal and spatial analyses, aimed at predicting future hurricane coverage and damage. 
We introduce three core models: the ARIMA time series prediction model for forecasting hurricane frequency, the K-Means clustering model for analyzing track distribution, and the Autoencoder model for simulating comprehensive hurricane coverage. 
We use the ARIMA model to forecast the frequency of future hurricane tracks based on historical data. Next, the K-means algorithm categorizes historical hurricane tracks to identify distribution patterns within specific clusters. Finally, the Autoencoder neural network simulates comprehensive hurricane coverage across the target region, with precise inputs and rigorous training mechanisms optimizing the model.

\subsection{Intensity of Tropical Cyclones}
\label{sec:4-1}

We use the North Atlantic basin's historical hurricane track data, HURricane DATa 2nd generation (HURDAT2)~\cite{fang2011smart}, provided by the National Hurricane Center (NHC). This database is a NOAA data set that offers comprehensive six-hourly information of tropical cyclones and subtropical cyclones in a comma-delimited, text file format. This information includes, but is not limited to, location, maximum winds, and central pressure of all known tropical cyclone events. This database records tropical cyclone activity in the Atlantic and East Pacific basins from 1851 to 2021. As a public resource, HURDAT2 provides data for analyzing trends, characteristics, and patterns of historical tropical cyclones, including hurricanes, typhoons, and tropical storms.

\textbf{Temporal-based Analysis.}
Hurricane occurrences in the North Atlantic basin exhibit temporal variations across both months and years. 
We first analyze the hurricane trend across different years.
Figure~\ref{fig:yearananlysis} illustrates the frequency of hurricane events from 1851 to 2021; here, the x-axis represents the historical years analyzed, while the y-axis corresponds to the frequency of hurricanes. A review of Figure~\ref{fig:yearananlysis} reveals that the frequency of hurricanes fluctuates among different years, with a discernible, albeit slight, upward trend across the tracking period. Then, we analyze the characteristics of hurricanes in each month.
The Atlantic hurricane season traditionally spans from June to November each year. While hurricanes may manifest in any month, the incidence is predominantly concentrated within this season, with September standing out as both the most frequent and the most damaging month for hurricanes. 


\textbf{Spatial-based Analysis.}
Hurricanes primarily form in tropical regions, where warm ocean waters supply the thermal energy required for initiation. In our study, we examine hurricane patterns through longitude and latitude, focusing on the start points of individual hurricane tracks. These starting locations reveal complex factors influencing hurricane formation, including sea surface temperatures and prevailing atmospheric conditions. Figure~\ref{fig:jointprobability} presents the joint probability density distributions of the origin of the hurricane by longitude and latitude. The top bar chart displays the marginal distribution of origins by longitude, while the right-hand bar chart represents the marginal distribution by latitude. Notably, hurricanes tend to concentrate around some places (such as 80$^\circ$ West on longitude and 13$^\circ$ North on latitude).

\subsection{Tropical Cyclone Occurrence}
\label{sec:4-2}

We begin our analysis by applying the Autoregressive Integrated Moving Average (ARIMA) model to predict hurricane frequency (i.e., the number of hurricane tracks). The ARIMA model~\cite{KOTU2019395}, a statistical-based numerical method, is commonly employed for time series analysis, particularly in forecasting univariate time series data.

ARIMA models are typically denoted by $ARIMA(p, d, q)$. Each of these parameters — $p$, $d$, and $q$ — are non-negative integers that represent distinct characteristics of the time series data and the model:
The parameter $p$ represents the order of the AutoRegressive (AR) component. It signifies the number of previous observations or lags that are taken into account when predicting the current observation. The parameter $d$ is the degree of differencing required to make the time series stationary. A time series is said to be stationary when its statistical properties (like mean and variance) do not change over time. Stationarity is essential for many time series forecasting methods, as it makes them more reliable. The parameter $q$ represents the order of the Moving Average (MA) component. It denotes how many past forecast error terms are used to predict the current observation. To determine the optimal values for these parameters, we conducted a grid search, and set up the parameters as $(4, 1, 1)$ in our study.

\renewcommand{\dblfloatpagefraction}{0.9}
\begin{figure*}[htbp]
    \centering
    \begin{minipage}[t]{0.5\textwidth}
        \centering
        \includegraphics[width=2.5in]{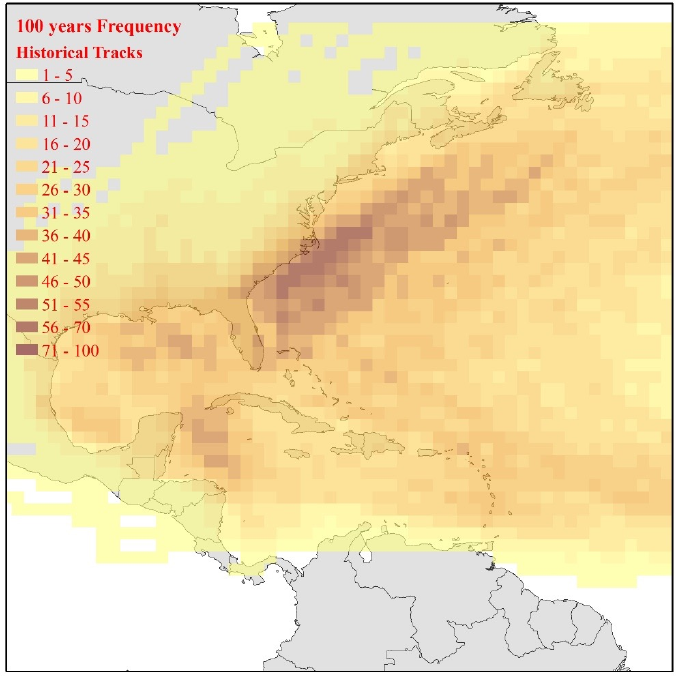}
        \\ 
        \textbf{(a)} Heat map of historical hurricane tracks for last 100 years
        \label{fig:historicaltrackplot-heat} 
    \end{minipage}%
    \hfill
    \begin{minipage}[t]{0.5\textwidth}
        \centering
        \includegraphics[width=2.5in]{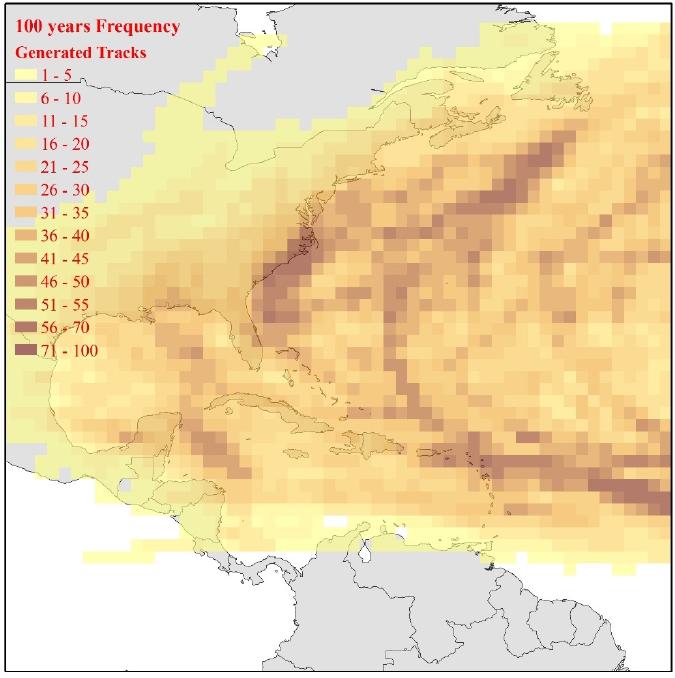}
        \\ 
        \textbf{(b)} Heat map of generated tracks of hurricane activity for 100 years using HurriCast
        \label{fig:generatedtrackplot-heat} 
    \end{minipage}

    \vspace{\floatsep} 
    \begin{minipage}[t]{0.5\textwidth}
        \centering
        \includegraphics[width=2.5in]{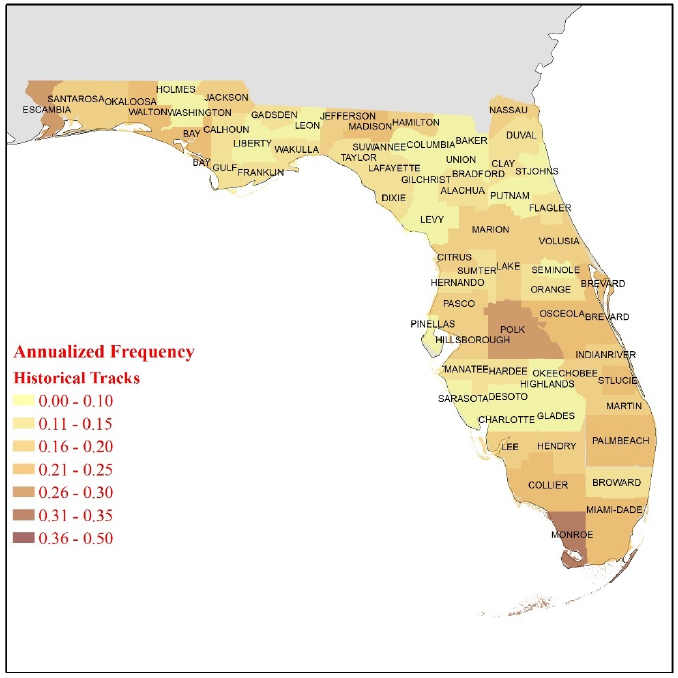}
        \\ 
        \textbf{(c)} Historical annualized frequency at Florida
        \label{fig:historicalheatplot-heat} 
    \end{minipage}%
    \hfill
    \begin{minipage}[t]{0.5\textwidth}
        \centering
        \includegraphics[width=2.5in]{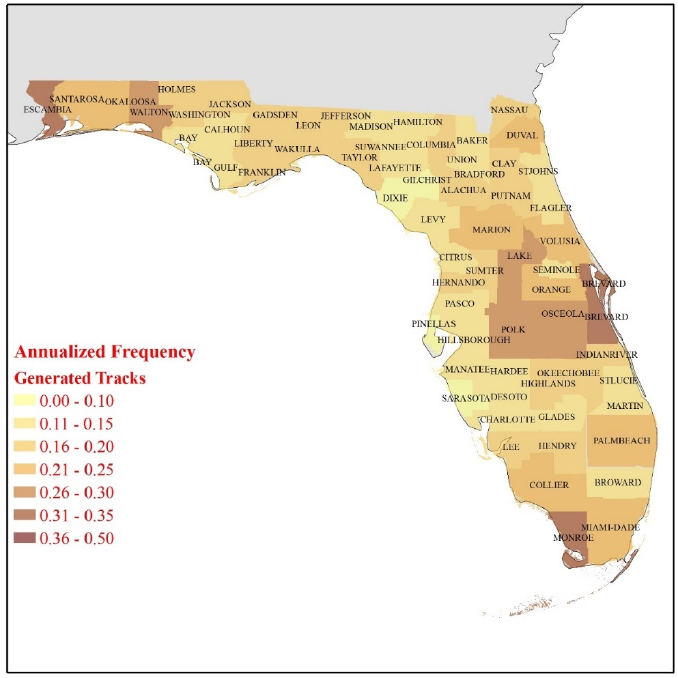}
        \\ 
        \textbf{(d)} Generated annualized frequency at Florida using HurriCast
        \label{fig:sampledtrackplot-heat} 
    \end{minipage}

    \caption{Comparison of the synthetic tracks (i.e., HurriCast) and historical tracks (i.e., the ground-truth) form frequency.}
    \label{autoencoder-perfomance}
    \vspace{-10pt}
\end{figure*}
\subsection{Start Points of Tropical Cyclones}
\label{sec:4-3}

Given the estimated hurricane frequency, our next step is to project how hurricane tracks will be distributed across a specific target region. To achieve this, we employ a sampling-based approach to figure out the number of hurricanes at every grid point.

In the first step, we use the K-means algorithm to examine historical hurricane tracks from the target region. This entails grouping hurricane tracks with similar movement patterns into separate clusters. 

On the second step, we analyze how the distribution of hurricane paths varies within each individual cluster. Then, based on the estimated hurricane frequency, we perform a sampling process to estimate the frequency distribution on each cluster (the number of hurricane tracks per cluster).

For the third step, we systematically go through each cluster in historical dataset, selectively extracting typical hurricane tracks with the corresponding proportion. 

This approach helps us create a subset of hurricane starting points and hurricane tracks that genuinely represent instances for each cluster. These starting points play a pivotal role in determining the hurricane number within each cluster, while the extracted hurricane tracks serve as seeds for estimating hurricane coverage in the following process - an Autoencoder model that will be elaborated.

\begin{figure*}[htbp]
    \centering
    \begin{minipage}[t]{0.3\textwidth}
        \centering
        \includegraphics[width=2in]{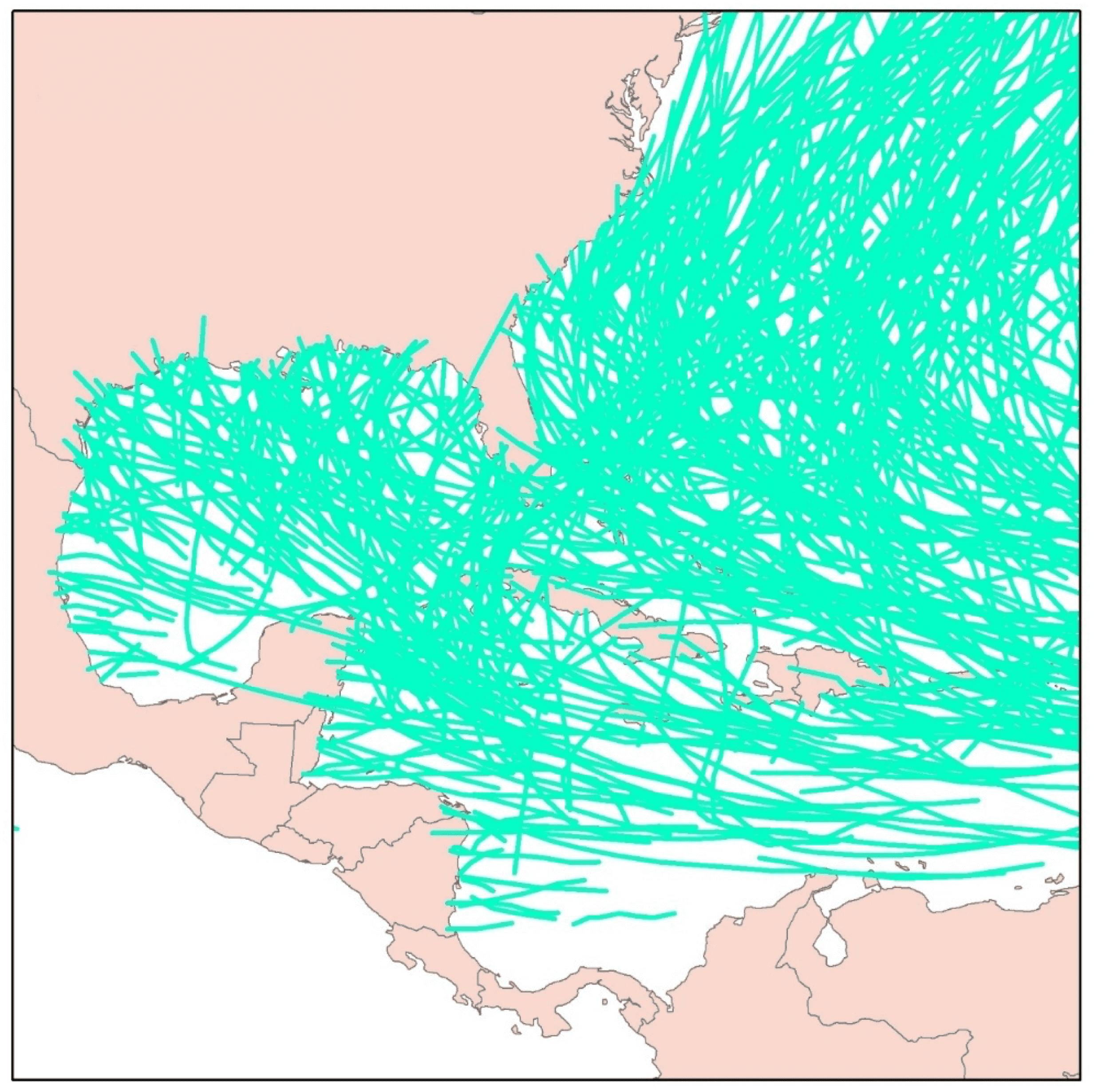}
        \\ 
        \textbf{(a)} Historical tropical cyclone (CAT$\geq$2)
        \label{fig:historicaltrackplot} 
    \end{minipage}%
    \hfill
    \begin{minipage}[t]{0.3\textwidth}
        \centering
        \includegraphics[width=2in]{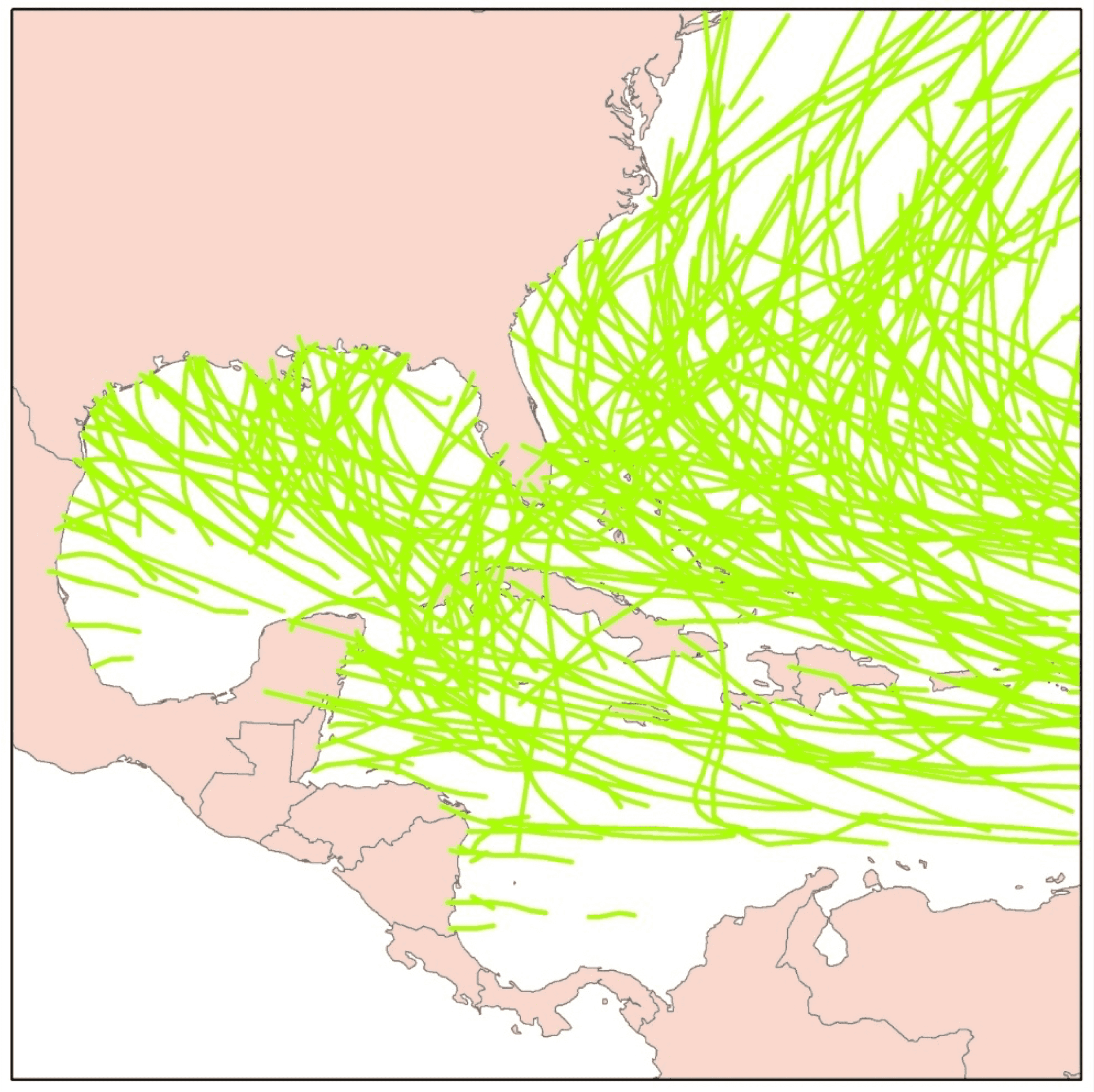}
        \\ 
        \textbf{(b)} Historical tropical cyclone (CAT$\geq$3)
        \label{fig:generatedtrackplot} 
    \end{minipage}%
    \hfill
    \begin{minipage}[t]{0.3\textwidth}
        \centering
        \includegraphics[width=2in]{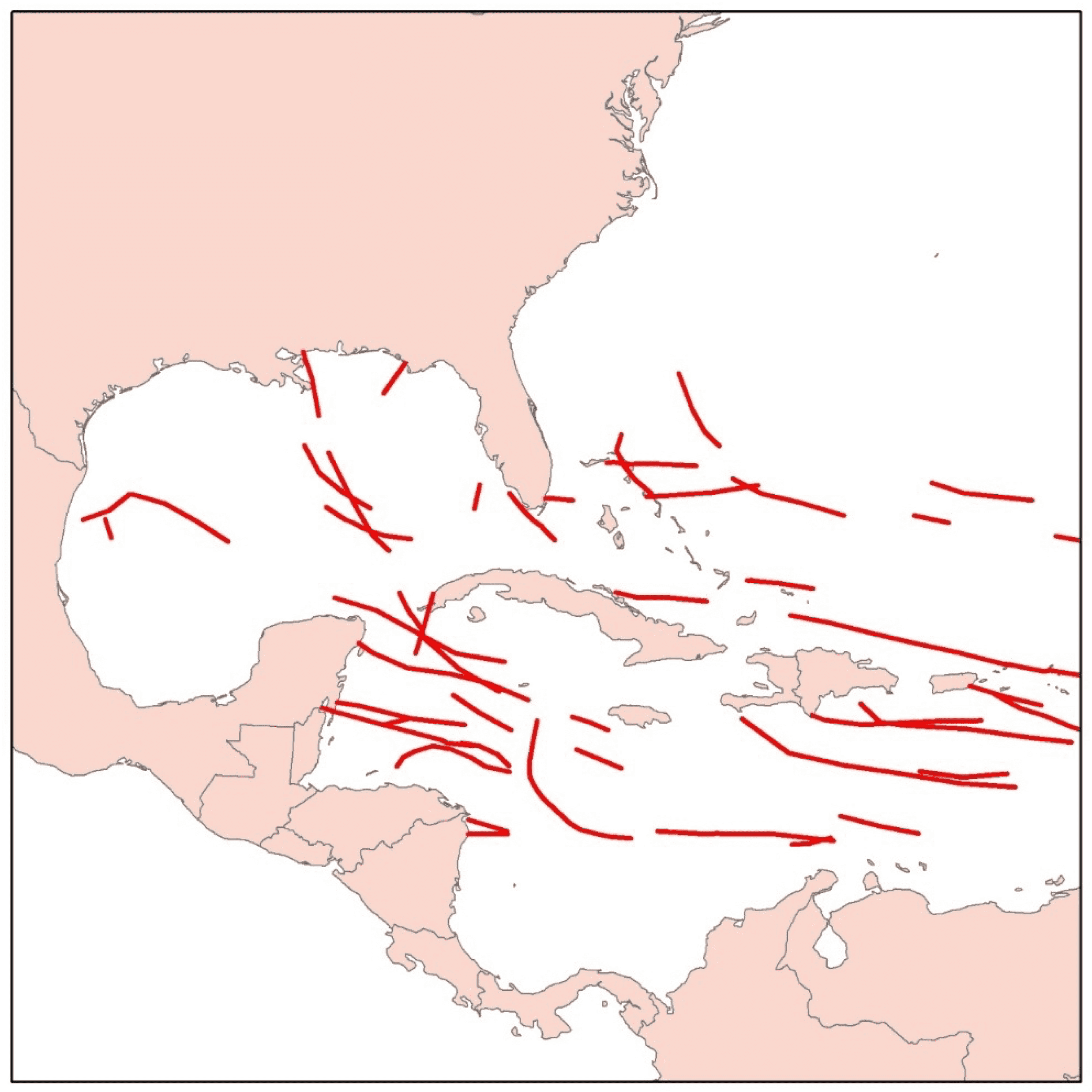}
        \\ 
        \textbf{(c)} Historical tropical cyclone (CAT$\geq$5)
        \label{fig:historicaltrackplot2} 
    \end{minipage}

    \vspace{\floatsep} 
    \begin{minipage}[t]{0.3\textwidth}
        \centering
        \includegraphics[width=2in]{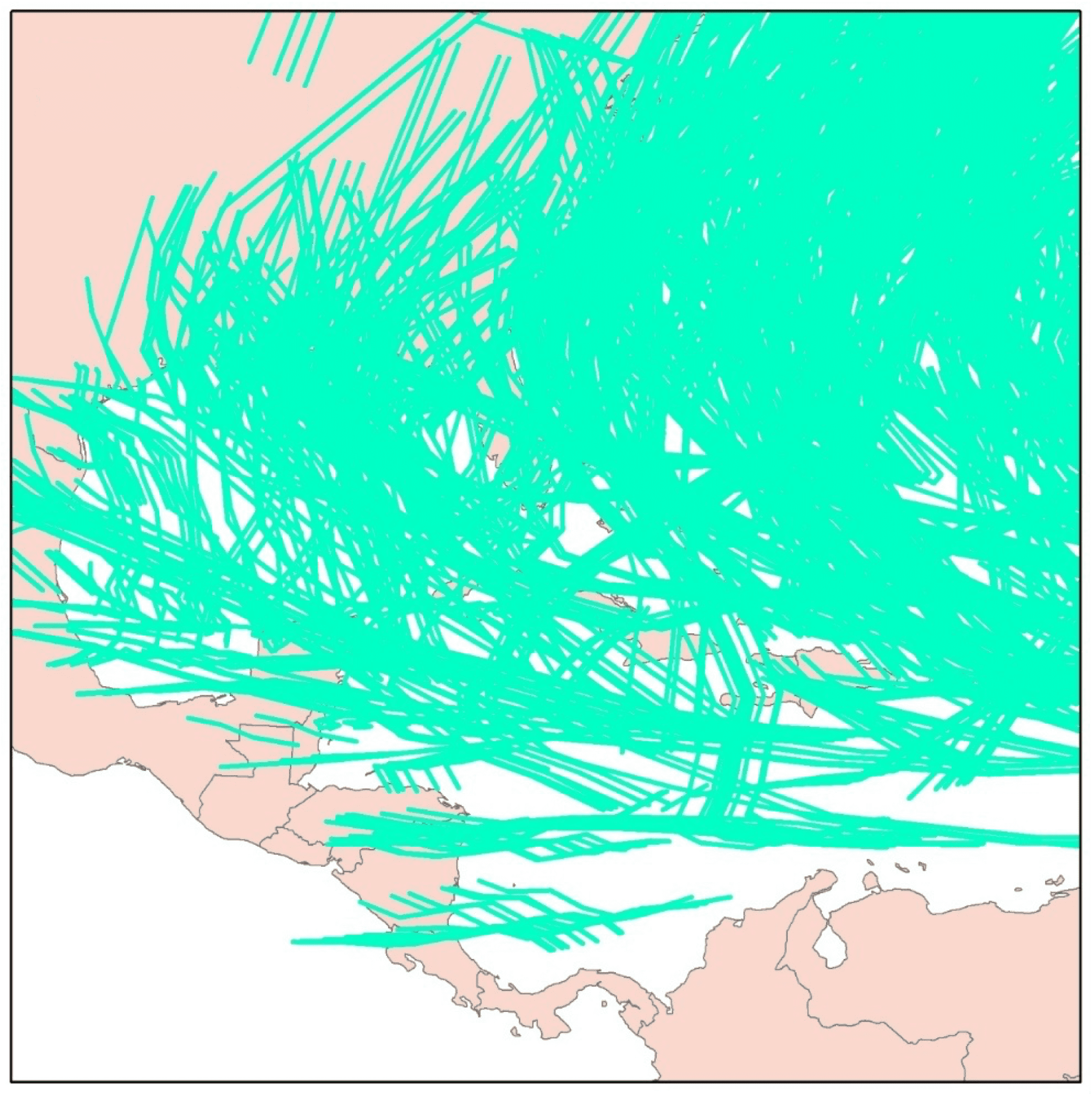}
        \\ 
        \textbf{(d)} Generated tropical cyclone (CAT$\geq$2)
        \label{fig:historicalheatplot} 
    \end{minipage}%
    \hfill
    \begin{minipage}[t]{0.3\textwidth}
        \centering
        \includegraphics[width=2in]{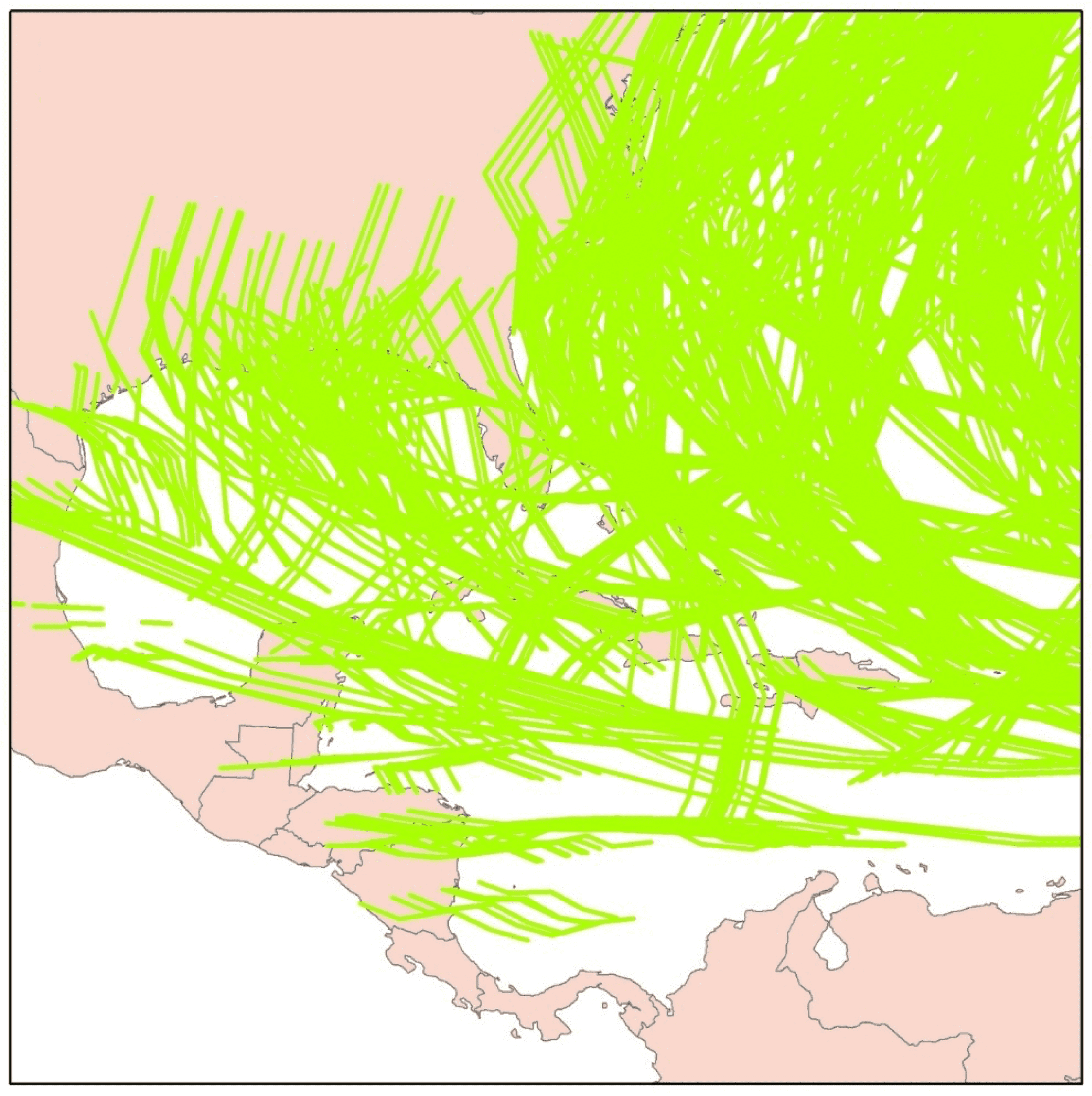}
        \\ 
        \textbf{(e)} Generated tropical cyclone (CAT$\geq$3)
        \label{fig:sampledtrackplot} 
    \end{minipage}%
    \hfill
    \begin{minipage}[t]{0.3\textwidth}
        \centering
        \includegraphics[width=2in]{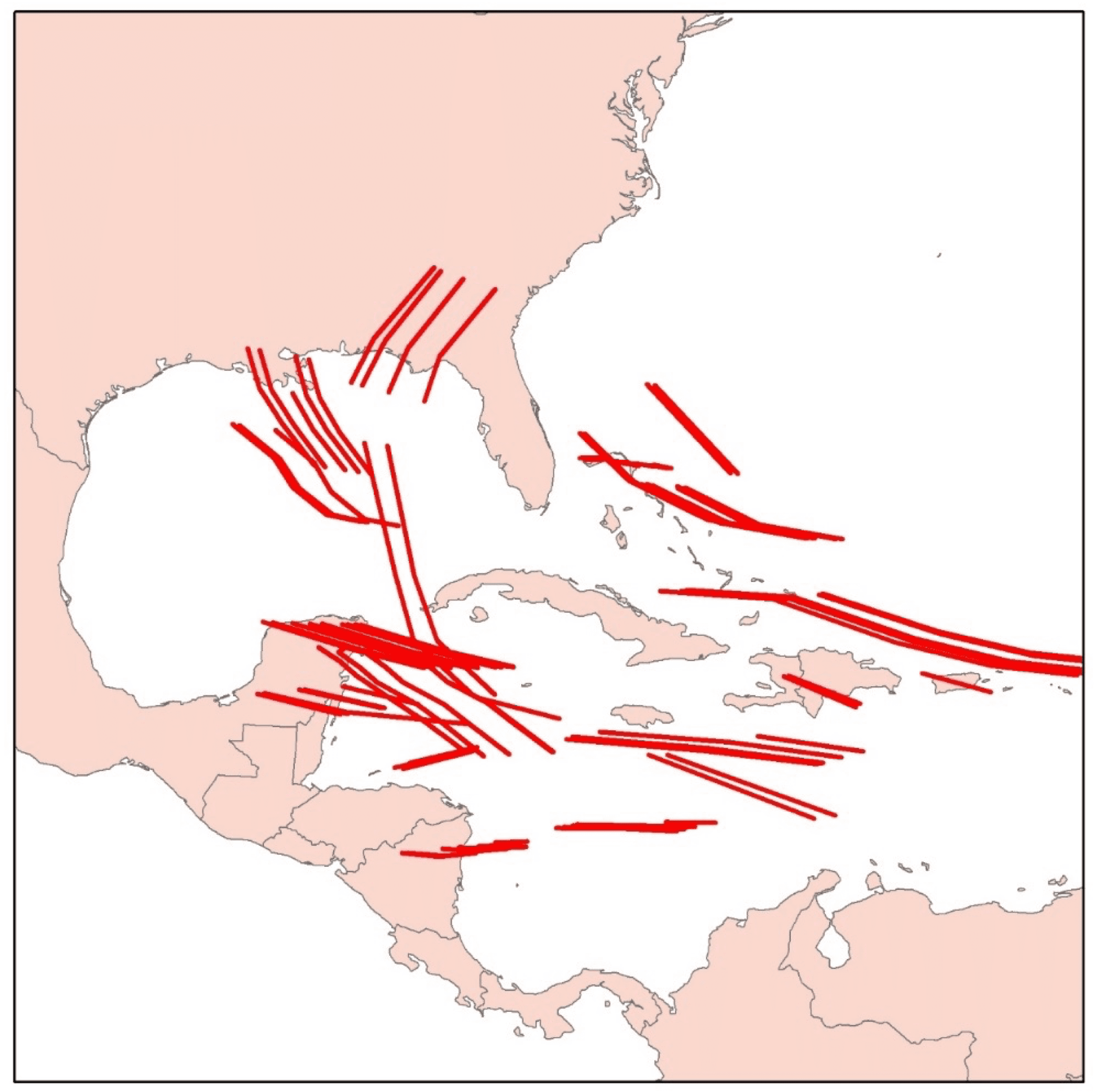}
        \\ 
        \textbf{(f)} Generated tropical cyclone (CAT$\geq$5)
        \label{fig:sampledtrackplot2} 
    \end{minipage}

    \caption{Comparison of different categories (CAT) for generated and historical hurricane tropical cyclone.}
    \label{cat-comparison}
    \vspace{-10pt}
\end{figure*}

\subsection{The Coverage of Tropical Cyclones }
\label{sec:4-4}
Based on the hurricane distribution per cluster and their corresponding seeds (i.e., the associated hurricane tracks), we leverage an Autoencoder model to replicate hurricane coverage across the target region.

The Autoencoder~\cite{hinton2006reducing}, a widely utilized neural network model, adeptly generates a compact representation of input data. This architecture comprises two essential components: an encoder and a decoder. The encoder transforms input data into a latent space, while the decoder reconstitutes the original data from this latent representation. 
 
In this study, we introduce \textit{a perturbation module} into the encoder that introduces Gaussian noise to the latent features. This step introduces an element of randomness, simulating the variability in hurricane occurrences. Notably, it's important to recognize that simulating the detailed process of hurricane generation using numerical methods is intricate and can be challenging to achieve optimal accuracy~\cite{jamous2023physics}.
In our approach, our objective is to provide an estimation of hurricane coverage for insurance purposes. We adopt a more straightforward and lightweight methodology, employing historical data as seeds and introducing Gaussian noise to simulate the stochastic nature of hurricane occurrences.
 
Specifically, both the encoder and decoder consist of four fully connected layers augmented by two batch normalization layers. In the disturbance addition module, we use random multiplicative noise, we multiply the features output by the encoder by a positive number (between 0 and 1) that obeys a random normal distribution and then feed the transformed features into the decoder. At last, the decoder outputs new hurricane tracks. 

For the input data, we choose an alternative approach to using raw data from HURDAT2. Instead, we employ spatial interpolation on hurricane trajectories, resulting in uniformly spaced trajectory points. These points undergo normalization to ensure consistent scaling, subsequently serving as the input data. This interpolation procedure plays a pivotal role in crafting a well-organized and evenly dispersed dataset, thereby elevating the quality of subsequent analyses. Post-interpolation, each hurricane trajectory is condensed into a set of 20 data points, each characterized by three dimensions: latitude, longitude, and minimum pressure. 

Throughout the model training phase, our choice of loss function is the Mean Squared Error (MSE), with optimization performed by the Adam optimizer utilizing a learning rate of 1e-3. Also, to counteract overfitting during training, we periodically introduce random Gaussian noise.

\subsection{Evaluation}

\textbf{Experimental setup.} To assess the efficacy of HurriCast, we utilized historical data from 1875 to 2010 from HURDAT2 as our training dataset. This dataset comprises 1134 hurricane tracks. Each track encompasses several track points, with each point recorded at 6-hour intervals. These points detail the hurricane's longitude, latitude, and intensity.
Utilizing this dataset, we predicted the hurricane coverage over a 10-year span, from 2011 to 2021. The historical data from this period served as the ground-truth to assess the performance of HurriCast. Within HurriCast, Pytorch~\cite{paszke2019pytorch}was employed for model training and inference.

To better quantitatively compare the hurricane coverage statistics obtained from the synthetic tracks with those from the historical tracks, we calculated the statistics from the synthetic tracks and presented them using a heat map. Figure ~\ref{autoencoder-perfomance} evaluates HurriCast's ability to reproduce historical North Atlantic tropical cyclone frequencies through comparisons of 100-year basin-wide frequency maps (Fig. 7a,b) and annualized frequencies impacting Florida (Fig. 7c,d). The HurriCast-generated frequency heat map (Fig. 7b) exhibits a high degree of agreement with the historical heat map (Fig. 7a), capturing the peak frequency corridor extending from the mid-Atlantic through the Caribbean and Gulf of Mexico. HurriCast also skillfully replicates observed cyclone frequencies over Florida (Fig. 7d), with a peak annualized frequency of 0.36-0.50 cyclones per year in the Florida Keys, closely matching the historical maximum of 0.36-0.50 (Fig. 7c). These results demonstrate that HurriCast can generate synthetic hurricane climatologies that are highly consistent with observed frequencies on both basin-wide and regional scales relevant for landfall risk assessment.

Figure~\ref{cat-comparison} compares synthetic tropical cyclone tracks generated by HurriCast to historical tropical cyclone tracks for three intensity categories: tropical storms and hurricanes (CAT$\geq$2), major hurricanes (CAT$\geq$3), and Category 5 hurricanes (CAT$\geq$5). The spatial distribution and geometries of the HurriCast tracks (Fig. 7d-f) display remarkable similarity with the historical record (Fig. 7a-c), particularly for the most intense cyclones. HurriCast generates synthetic Category 5 tracks (Fig. 7f) that closely match the observed clustering and trajectories of historical Category 5 tracks (Fig. 7c). However, the HurriCast tracks noticeably extend over continental land masses, especially for the lower intensity categories (Fig. 7d,e), while historical tracks rapidly dissipate after landfall due to the loss of the warm ocean as an energy source. This suggests that future refinements of HurriCast should account for land interaction to yield more physically realistic track life cycles. Nonetheless, these results demonstrate HurriCast's ability to generate realistic tropical cyclone tracks across the intensity spectrum that are consistent with historical climatology.

Several general qualitative results can be observed:
(1) The spatial trends along the Atlantic basin are consistent. This suggests that HurriCast can accurately model the spatial trends of hurricane wind hazards.
(2) Differences exist between the wind hazards estimated from historical tracks and those from synthetic tracks. 
This suggests that the synthetic tracks exhibit less variability than the historical tracks from the past decade in the Atlantic basin. Given that the synthetic track model utilizes an extensive historical dataset, it naturally produces more conservative results.
(3) The severity of hurricane impacts, measured in terms of potential damage, exhibits a similar pattern in both historical (Figure~\ref{cat-comparison}a, b and c) and synthetic data (Figure~\ref{cat-comparison} d, e, and f).

\section{Summary and Future Work}
Future refinements of HurriCast should focus on improving the model's representation of land interactions, particularly for lower-intensity storms, to more accurately simulate the rapid dissipation of tracks after landfall and produce more physically realistic track life cycles. Moreover, real-time hurricane prediction is a critical area for further development, various methods, such as Generative Adversarial Networks (GANs) and Long-Short-Term Memory (LSTM) networks could be adopted in this problem.

\section{Acknowledgements}

This research is supported by U.S. National Science Foundation OAC-2417849: CyberTraining: EcoTern: Pioneering a CI Workforce for Sustainable and Transdisciplinary Environmental Science Research and IIS-2331908: PARTNER: An AI/ML Collaborative for Southeast Florida Coastal Environmental Data and Modeling Center.

\bibliography{main}

\end{document}